\documentclass[10pt]{article} 
\usepackage[preprint]{tmlr}

\usepackage{hyperref}
\usepackage{url}

\usepackage{amsmath}
\usepackage{amssymb}
\usepackage{booktabs}
\usepackage{comment}

\usepackage{pgfplots,pgfplotstable}
\DeclareUnicodeCharacter{2212}{?}
\usepgfplotslibrary{groupplots,dateplot}
\usetikzlibrary{patterns,shapes.arrows}
\pgfplotsset{compat=newest}
\usepackage{subcaption}

\title{Transfer Learning for Segmentation Problems: \\Choose the Right Encoder and Skip the Decoder}

\author{\name Jonas Dippel
        \email j.dippel@campus.tu-berlin.de \\
        \addr Technische Universität Berlin \\
        \addr Bayer AG, M\"ullerstr. 178, 13353 Berlin, Germany
        \AND
        \name Matthias Lenga
        \email matthias.lenga@bayer.com \\
        \addr Bayer AG, M\"ullerstr. 178, 13353 Berlin, Germany
        \AND
        \name Thomas Goerttler
        \email thomas.goerttler@tu-berlin.de \\
        \addr Technische Universität Berlin, Chair of Neural Information Processing
        \AND
        \name Klaus Obermayer
        \email klaus.obermayer@tu-berlin.de \\
        \addr Technische Universität Berlin, Chair of Neural Information Processing \\
        Bernstein Center for Computational Neuroscience Berlin
        \AND
        \name Johannes Höhne \email johannes.hoehne@bayer.com \\
        \addr Bayer AG, M\"ullerstr. 178, 13353 Berlin, Germany}

\def\month{MM}  
\def\year{YYYY} 
\def\openreview{\url{https://openreview.net/forum?id=XXXX}} 

\begin{document}

\maketitle

\begin{abstract}
It is common practice to reuse models initially trained on different data to increase downstream task performance. Especially in the computer vision domain, ImageNet-pretrained weights have been successfully used for various tasks. In this work, we investigate the impact of transfer learning for segmentation problems, being pixel-wise classification problems that can be tackled with encoder-decoder architectures. We find that transfer learning the decoder does not help downstream segmentation tasks, while transfer learning the encoder is truly beneficial. 
We demonstrate that pretrained weights for a decoder may yield faster convergence, but they do not improve the overall model performance as one can obtain equivalent results with randomly initialized decoders.
However, we show that it is more effective to reuse encoder weights trained on a segmentation or reconstruction task than reusing encoder weights trained on classification tasks. This finding implicates that using ImageNet-pretrained encoders for downstream segmentation problems is suboptimal. We also propose a contrastive self-supervised approach with multiple self-reconstruction tasks, which provides encoders that are suitable for transfer learning in segmentation problems in the absence of segmentation labels.
\end{abstract}

\section{Introduction}
Transfer learning and reusing pretrained weights is common practice when training deep learning models. Reusing weights from a model that was pretrained on a large scale dataset often has several advantages: faster training time, reduced costs, ecological footprint and improved performance - especially in low-data regimes.
For computer vision problems, the use of ImageNet-pretrained weights for model initialization has de facto become standard practice as these encode information related to the visual content of millions of images from diverse domains. To better leverage the underlying structure in the data, advanced self-supervised learning methods such as SimCLR \citep{chen2020simple}, BYOL \citep{grill2020bootstrap}, and MoCo \citep{he2020momentum} have been proposed in the last years. Common to all approaches is that they only pretrain an encoder network.
In order to capture fine-grained visual features in the latent representations, the ConRec framework \citep{dippel2021towards} extends upon SimCLR by incorporating a decoder network and jointly optimizing a contrastive and a self-reconstruction loss.

\begin{figure}[ht]
	\centering
	\includegraphics[width=0.9\linewidth, trim=0cm 7cm 0cm 0cm]{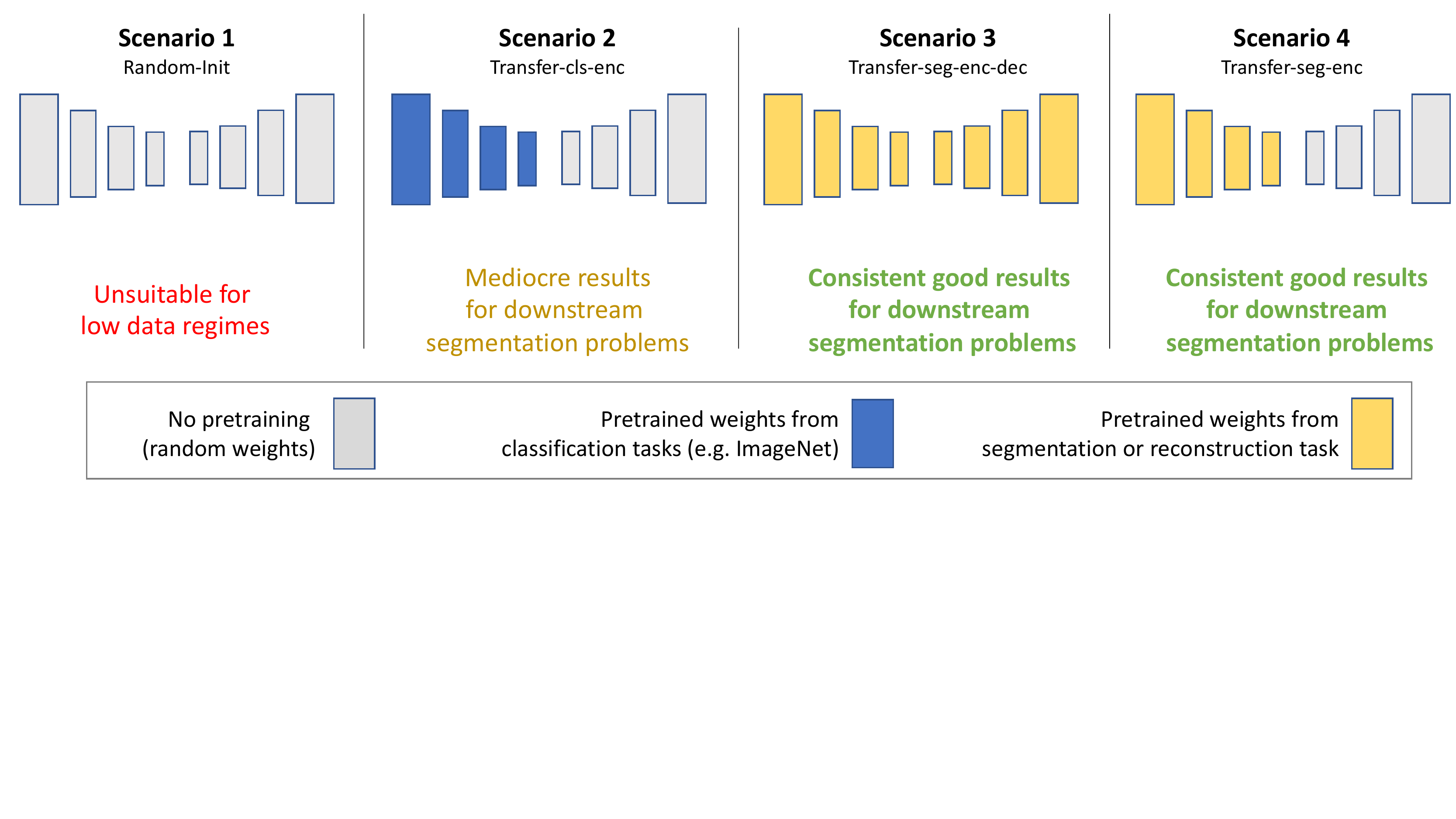}
	\caption{Transfer learning scenarios for segmentation problems. Note that Scenario 2 is commonly applied in literature and our experimental results show that this is suboptimal.}
	\label{fig:transfer_learning_approaches}
\end{figure}
For segmentation tasks, encoder-decoder architectures often reuse ImageNet weights as an initialization for the encoder, whereas the initial weights of the decoder are commonly initialized by drawing from a random distribution: \cite{minaee2021image} surveyed more than 100 recent image segmentation algorithms, and they state that "many people use a model trained on ImageNet" as the encoder part of the network, and re-train their model from those initial weights.
In the context of medical image segmentation, it has been observed that  self-supervised pretraining with a reconstruction task has the potential to significantly improve the model performance on downstream segmentation tasks  \citep{zhou2019models,haghighi2020learning}.

It is yet not well understood, though, which underlying factors are driving the reported improvements. 

This study aims at systematically analyzing the effect of different pretraining and initialization approaches for encoder-decoder networks on downstream segmentation tasks. Hereby we limit our experimental setup to encoder-decoder architectures as they are most prominently applied for image segmentation.
The competing initialization approaches comprise (a) encoder pretrained on a pretext classification task, (b) entire network pretrained on a pretext reconstruction or segmentation task, and (c) encoder pretrained on a pretext reconstruction/segmentation task with randomly initialized decoder -- see Fig.~\ref{fig:transfer_learning_approaches}. Clearly, these approaches require different levels of annotations, i.e., no labels, class labels, or segmentation masks, which has direct implications on their practical applicability.
A comprehensive quantitative analysis of the different learning scenarios is provided for image segmentation tasks of varying sizes derived from four distinct data sources. To simplify and focus our analysis, we limit our experiments to the well-known U-Net architecture \citep{ronneberger2015u} which is widely used in medical segmentation tasks.

Our experiments confirm the intuition that it is most beneficial to pretrain on pretext tasks that are of the same type as the downstream task. Hence, choosing by default an encoder that was simply pretrained on ImageNet classification as initialization may, in general, be suboptimal for downstream image segmentation problems. Interestingly, we observe that in our experimental setup, the pretraining of the decoder seems to be only of minor importance. Thus, it may be of key importance to  carefully select encoder weights that are used for initialization, whereas the decoder branch can be easily learned from scratch on the downstream task. 
Next to the quantitative analysis of different transfer learning scenarios for segmentation problems, we also provide a qualitative analysis that further explains the observed patterns. We compare encoder representations from different models, and we find that an encoder which is trained with a segmentation pretext task provides fundamentally different representations than an encoder from a classification pretext task.

\section{Evaluation Environment}
Below, we provide a detailed description of the evaluation environment that was used to investigate several transfer learning scenarios.

\subsection{Datasets}
\begin{figure}[h!]
	\centering
	\begin{subfigure}[b]{0.3\linewidth}
		\centering
		\includegraphics[width=\linewidth]{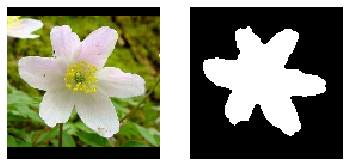}
		\caption{Oxford Flowers 17}
	\end{subfigure}
	\begin{subfigure}[b]{0.3\linewidth}
		\centering
		\includegraphics[width=\linewidth]{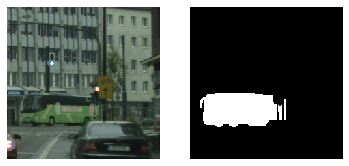}
		\caption{Cityscape Bus}
	\end{subfigure}
	\begin{subfigure}[b]{0.3\linewidth}
		\centering
		\includegraphics[width=\linewidth]{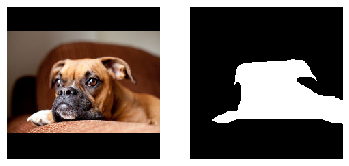}
		\caption{Oxford IIIT Pets}
	\end{subfigure}
	\par\bigskip
	\begin{subfigure}[b]{0.3\linewidth}
		\centering
		\includegraphics[width=\linewidth]{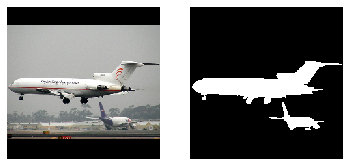}
		\caption{Pascal VOC airplane}
	\end{subfigure}
	\begin{subfigure}[b]{0.3\linewidth}
		\centering
		\includegraphics[width=\linewidth]{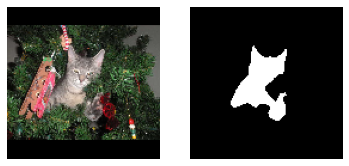}
		\caption{Pascal VOC cat}
	\end{subfigure}
	\begin{subfigure}[b]{0.3\linewidth}
		\centering
		\includegraphics[width=\linewidth]{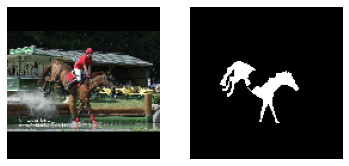}
		\caption{Pascal VOC horse}
	\end{subfigure}
	\hfill
	\caption{Samples from the six segmentation datasets:  Flowers, Cityscape, Oxford IIIT Pets and Pascal VOC airplane, cat, horse with their respective annotations.}
	\label{fig:seg-samples}
\end{figure}

We used six benchmark datasets:  Flower Segmentation \citep{Nilsback08}, Cityscapes \citep{Cordts2016Cityscapes}, Oxford IIIT Pets \citep{parkhi12a} and three subsets of the Pascal VOC dataset (airplanes, cats, horse) \citep{pascal-voc-2012} -- see also Fig.~~\ref{fig:seg-samples} and Table \ref{tab:dataset} for further details. As those tasks are rather easy, we vary the amount of training data and observe the effect during evaluation. We vary the fraction samples used for training (1\%, 5\%, 20\%) for all datasets and use the remaining samples for evaluation.

\begin{table}[h!]
    \small
	\centering
	\caption{Dataset size of our six benchmark datasets.}
	\label{tab:dataset}
	\begin{tabular}{ccccccc}
	\toprule
	Dataset & Flowers 17 & Oxford Pets & Cityscape Bus & VOC airplane & VOC cat & VOC horse \\
	\midrule
	\#Samples & 849 & 7375 & 483 & 178 & 250 & 147 \\
	\bottomrule
\end{tabular}
\end{table}

\subsection{Related work -- SimCLR, Reconstruction and ConRec}
\label{sec:conrec}

We want to compare popular self-supervised learning methods such as the SimCLR framework \citep{chen2020simple,chen2020big}, MoCo \citep{he2020momentum} and BYOL \citep{richemond2020byol} that only train an encoder with methods that also train a decoder. SimCLR learns representations by learning that views from the same image are similar and that views from different images are dissimilar. 
Reconstruction-based methods commonly train a model to predict the content of artificially masked  parts of an image by optimizing a $l_2$ reconstruction loss \citep{pathak2016context,zhou2019models,dippel2021towards}. To overcome information redundancy in images, recent approaches suggest masking a very high portion of random patches  in order to create a self-supervised task for learning expressive visual features \citep{he2021masked, bao2021beit,zhou2021ibot, bachmann2022multimae}.
The ConRec framework \citep{dippel2021towards} combines the contrastive task of the SimCLR framework with a reconstruction task to initialize the decoder.
Classification networks tend to extract features which are invariant within a certain class while reconstruction networks extract features capable of capturing fain-grained visual details of a given instance.
In order to balance intra-class invariance and feature richness, the ConRec framework jointly optimizes a contrastive and a reconstruction loss.
Several recent studies have been following a similar approach \citep{li2020making, zhou2021preservational}, to tackle "feature suppression" of fine-grained details.

The vanilla ConRec model as described in \cite{dippel2021towards} has shown promising results for classification tasks. Segmentation tasks require pixel-level precision and therefore it is necessary to exploit fine-grained visual details from an image.

\begin{figure}[h!]
	\centering
	\includegraphics[width=.8\linewidth]{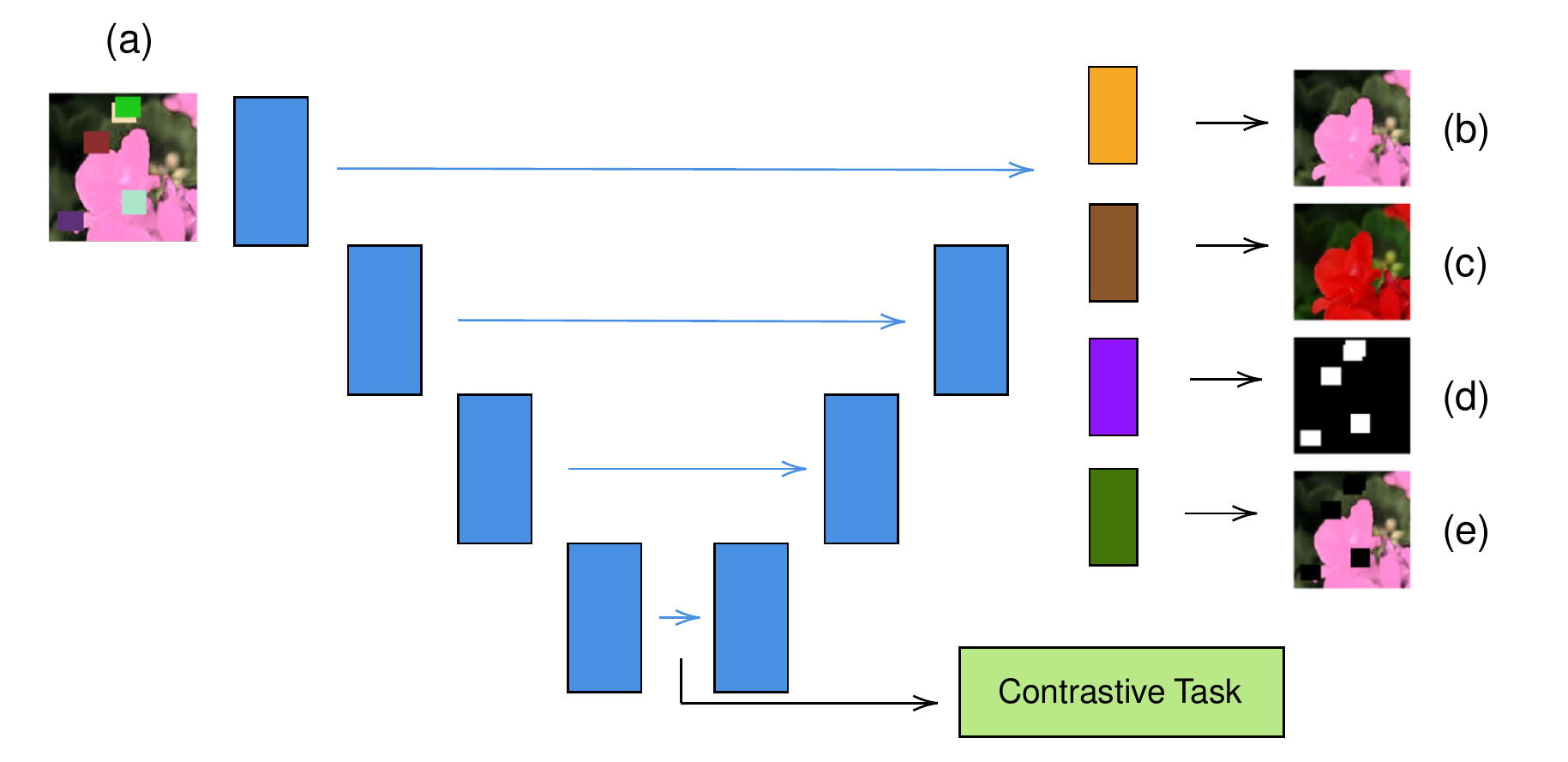}
	\caption{
	Four reconstruction tasks that are performed simultaneously when training the ConRec framework. Given the input image (a), the ConRec model outputs the images (b)-(e) with its four reconstruction heads.}
	\label{fig:multi-reconstruction}
\end{figure}

In order to tailor the ConRec approach towards segmentation problems and to build more robust and generalized representations in the decoder, we extent this concept: instead of a single reconstruction task, we challenge the ConRec model with multiple reconstruction tasks while sharing most of the weights between the different tasks. The four tasks include the normal reconstruction of the masked image, segmenting the locations where masks are applied to the image, and reverting the image from the color jittered colorspace to the original color space. Given one example image, the four target images for each reconstruction task are shown in Fig.~\ref{fig:multi-reconstruction}. Subfigure (a) shows an example input image and (b) - (e) show the respective reconstruction targets. (b) depicts the normal reconstruction of the image, (c) also includes the projection to the original color space, (d) highlights the regions in the image where a mask can be found, and (e) fills the masks with black color. The first three blocks of the decoder are shared between the reconstruction tasks, and the last decoder block is unique to each task.

\begin{table*}[h!]
    \small
	\centering
	\caption{Finetuning results for various pretrained models and scenarios on the Oxford Flowers 17 and Oxford Pets datasets. Displayed values are the dice coefficient on the evaluation set.}
	\label{tab:flower-pets}
	\begin{tabular}{cccccccc}
	\toprule
      \multicolumn{2}{c}{}    &          \multicolumn{3}{c}{Flowers}                    &   \multicolumn{3}{c}{Pets}            \\
      \cmidrule(lr){3-5}\cmidrule(lr){6-8}
 Scenario & Model &       1\% (8) &                  5\% (42) &            20\% (170) &          1\% (74) &          5\% (368) &           20\% (1475) \\
\midrule
1 (random-init) &                   Random &  79.1 $\pm$ 1.0 &            89.4 $\pm$ 0.8 &  94.3 $\pm$ 0.0 &  71.7 $\pm$ 5.2 &   84.2 $\pm$ 0.9 &  88.2 $\pm$ 0.3 \\
2  (cls-enc) &                   SimCLR &  77.7 $\pm$ 0.4 &            88.8 $\pm$ 0.3 &  94.1 $\pm$ 0.1 &  79.2 $\pm$ 1.1 &   85.1 $\pm$ 0.1 &  88.2 $\pm$ 0.4 \\
3   (seg-enc-dec) &           Reconstruction &  84.4 $\pm$ 0.6 &            91.8 $\pm$ 0.2 &  95.0 $\pm$ 0.1 &  80.5 $\pm$ 1.5 &   86.8 $\pm$ 0.2 &  89.6 $\pm$ 0.3 \\
3 (seg-enc-dec) &             ConRec &  85.7 $\pm$ 1.8 &            92.2 $\pm$ 0.4 &  95.3 $\pm$ 0.1 &  83.1 $\pm$ 0.5 &   87.5 $\pm$ 0.6 &  90.0 $\pm$ 0.1 \\
4   (seg-enc) &   Reconstruction &  84.6 $\pm$ 0.2 &            92.0 $\pm$ 0.1 &  95.0 $\pm$ 0.0 &  81.2 $\pm$ 0.1 &   86.9 $\pm$ 0.5 &  89.5 $\pm$ 0.1 \\
4  (seg-enc)  &      ConRec &  84.7 $\pm$ 0.6 &            92.1 $\pm$ 0.1 &  95.3 $\pm$ 0.1 &  83.3 $\pm$ 0.6 &   87.2 $\pm$ 0.4 &  89.8 $\pm$ 0.2 \\
\bottomrule
\end{tabular}
\end{table*}

\subsection{Implementation Details}
As a preprocessing step, we resize the images with padding to the desired target size. During training, we randomly flip the image horizontally and take a random crop containing at least 50\% of the image while changing the aspect ratio maximally to ($\frac{3}{4}$, $\frac{4}{3}$) of the original image ratio. We use the dice loss as the objective function. Following the official ConRec implementation\footnote{\scriptsize{\url{https://github.com/bayer-science-for-a-better-life/contrastive-reconstruction}}}, our U-Net has four encoder and decoder blocks consisting of two convolutional layers with batch normalization and max pooling/upsampling in the encoder/decoder. This results in a total of 8.65M parameters for the model. While training for a specific subclass from the Pascal VOC or Cityscapes dataset, we use all the images where there is at least one pixel of the target class visible and discard all other images. 


\section{Do we need a Pretrained Decoder?}\label{sec:pretrain_decoder}

We evaluate various model training scenarios in order to assess whether or not a pretrained decoder is advantageous for downstream segmentation tasks. Note that the very same model encoder-decoder architecture is used for all model training scenarios, we only alter the weight initialization. For pretraining the self-supervised models, we use the Oxford Flowers 102 and Oxford Pets dataset. For finetuning, we train and evaluate models for two segmentation datasets (Oxford Flowers 17; Oxford Pets) with different amount of data (1\%, 5\% and 20\%) used for training. Comprehensive results are listed in Table \ref{tab:flower-pets}. As expected, transfer learning helps in general, hence scenario 1 with all weights initialized randomly yields the worst results.

Using ConRec as pretext task outperforms all other methods across the different data regimes and datasets, highlighting that the multiple reconstruction tasks lead to more general representations which can be generalized in downstream segmentation tasks. It slightly outperforms the models that were initialized with the reconstruction pretext task. With 20 \% training data ($\approx$170 training samples), the differences in performance are marginal, and the dice coefficient is high. This is because the segmentation task is relatively easy as the objects appear prominent in the image and sufficient training data is available to overrule the model initialization.

We aim to investigate if the observed performance increases in the low-data regime are the result of using a pretrained decoder. Therefore, we randomize the decoder of the models that were pretrained with a reconstruction task (i.e.~scenario 4) and compare the resulting performance to the fully initialized model (i.e.~scenario 3).  The results show that randomizing the decoder does not have a significant influence on the final performance. This finding was further validated on a real-world dataset -- see Section \ref{sec:suppl_realworld_data}.

\begin{figure*}[h!]
	\centering
	\begin{subfigure}[b]{0.49\linewidth}
		\centering
		\scalebox{0.9}{\begin{tikzpicture}

\definecolor{color0}{rgb}{0.12156862745098,0.466666666666667,0.705882352941177}
\definecolor{color1}{rgb}{1,0.498039215686275,0.0549019607843137}
\definecolor{color2}{rgb}{0.172549019607843,0.627450980392157,0.172549019607843}

\begin{axis}[
height=5cm,
legend cell align={left},
legend columns=3,
legend style={
  fill opacity=1,
  draw opacity=1,
  text opacity=1,
  at={(0.5,1.25)},
  anchor=north,
  draw=white!80!black,
  style={column sep=0.2cm}
},
tick align=outside,
tick pos=left,
width=8cm,
x grid style={white!69.0196078431373!black},
xlabel={Epoch},
xmajorgrids,
xmin=-0.5, xmax=208.5,
xtick style={color=black},
y grid style={white!69.0196078431373!black},
ylabel={Dice Coef},
ymajorgrids,
ymin=0.816555798053741, ymax=0.917143166065216,
ytick style={color=black}
]
\addplot [semithick, color0, mark=x, mark size=3, mark options={solid}]
table {%
9 0.896217167377472
19 0.905673921108246
29 0.90860116481781
39 0.908680319786072
49 0.909593343734741
59 0.910790860652924
69 0.909825384616852
79 0.909190475940704
89 0.911441087722778
99 0.910843074321747
109 0.911259770393372
119 0.911772549152374
129 0.91169548034668
139 0.911612331867218
149 0.912313997745514
159 0.911281526088715
169 0.911313831806183
179 0.912127256393433
189 0.912571012973785
199 0.910372376441956
};
\addlegendentry{ConRec}
\addplot [semithick, color1, mark=x, mark size=3, mark options={solid}]
table {%
9 0.855656445026398
19 0.869250297546387
29 0.882346749305725
39 0.885319352149963
49 0.891893744468689
59 0.899832904338837
69 0.903998672962189
79 0.902626216411591
89 0.905212461948395
99 0.908122539520264
109 0.906801342964172
119 0.905940890312195
129 0.908020794391632
139 0.91034734249115
149 0.91018533706665
159 0.910357356071472
169 0.910924077033997
179 0.911492586135864
189 0.911198735237122
199 0.911314427852631
};
\addlegendentry{ConRec RD}
\addplot [semithick, color2, mark=x, mark size=3, mark options={solid}]
table {%
9 0.821127951145172
19 0.839762926101685
29 0.853449285030365
39 0.856541454792023
49 0.866799294948578
59 0.872388184070587
69 0.871683716773987
79 0.876920163631439
89 0.880694031715393
99 0.879077196121216
109 0.883509516716003
119 0.883647441864014
129 0.878336131572723
139 0.884359300136566
149 0.88520073890686
159 0.884482562541962
169 0.886282086372375
179 0.884847223758698
189 0.885261297225952
199 0.885686814785004
};
\addlegendentry{SimCLR}
\end{axis}

\end{tikzpicture}}
		\caption{Flowers 17}
		\label{fig:flowers-decoder}
	\end{subfigure}
	\begin{subfigure}[b]{0.49\linewidth}
		\centering
		\scalebox{0.9}{\begin{tikzpicture}

\definecolor{color0}{rgb}{0.12156862745098,0.466666666666667,0.705882352941177}
\definecolor{color1}{rgb}{1,0.498039215686275,0.0549019607843137}
\definecolor{color2}{rgb}{0.172549019607843,0.627450980392157,0.172549019607843}
\definecolor{color3}{rgb}{0.83921568627451,0.152941176470588,0.156862745098039}

\begin{axis}[
height=5cm,
legend cell align={left},
legend columns=2,
legend style={
  fill opacity=1,
  draw opacity=1,
  text opacity=1,
  at={(0.5,1.35)},
  anchor=north,
  draw=white!80!black,
  style={column sep=0.2cm}
},
tick align=outside,
tick pos=left,
width=8cm,
x grid style={white!69.0196078431373!black},
xlabel={Epoch},
xmajorgrids,
xmin=-0.5, xmax=208.5,
xtick style={color=black},
y grid style={white!69.0196078431373!black},
ylabel={Dice Coef},
ymajorgrids,
ymin=0.479376070201397, ymax=0.840657688677311,
ytick style={color=black}
]
\addplot [semithick, color0, mark=x, mark size=3, mark options={solid}]
table {%
19 0.768976330757141
39 0.782566547393799
59 0.799732506275177
79 0.807013154029846
99 0.81143456697464
119 0.804109156131744
139 0.806680917739868
159 0.817336320877075
179 0.795554220676422
199 0.823730051517487
};
\addlegendentry{Segmentation}
\addplot [semithick, color1, mark=x, mark size=3, mark options={solid}]
table {%
19 0.545530736446381
39 0.617569208145142
59 0.665432929992676
79 0.730445086956024
99 0.776825368404388
119 0.7997887134552
139 0.803721129894257
159 0.813104033470154
179 0.824235796928406
199 0.824146747589111
};
\addlegendentry{Segmentation-EncOnly}
\addplot [semithick, color2, mark=x, mark size=3, mark options={solid}]
table {%
19 0.649967968463898
39 0.721601665019989
59 0.734398782253265
79 0.73738044500351
99 0.75335967540741
119 0.761210858821869
139 0.76499605178833
159 0.773019433021545
179 0.759884119033813
199 0.772546947002411
};
\addlegendentry{Classification}
\addplot [semithick, color3, mark=x, mark size=3, mark options={solid}]
table {%
19 0.665688812732697
39 0.63456130027771
59 0.702609717845917
79 0.714588940143585
99 0.714229702949524
119 0.713638961315155
139 0.72465991973877
159 0.725692629814148
179 0.739057302474976
199 0.725481450557709
};
\addlegendentry{Random}
\end{axis}

\end{tikzpicture}}
		\caption{Cityscape bus}
		\label{fig:dice_cityscapes_bus}
	\end{subfigure}
	\caption{Validation dice coefficient over the training time for two segmentation downstream tasks with different model initializations.}
	\label{fig:dice_over_epochs}
\end{figure*}
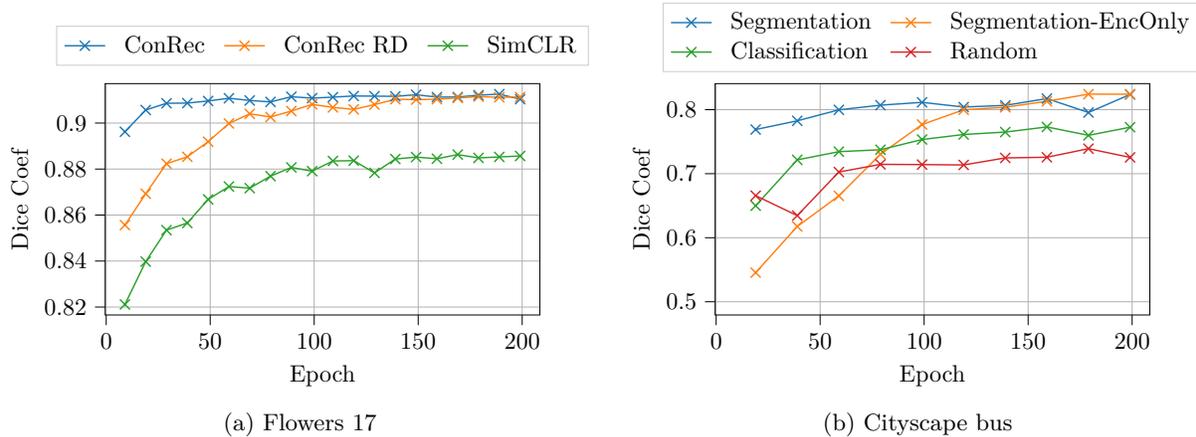

There are two potential explanations for this finding: Either the computational graph in a pretrained decoder is irrelevant for the downstream segmentation task, or even very little number of labeled training samples are required to restore the data transformations required. Fig.~\ref{fig:flowers-decoder} shows the dice coefficient on the test set for the 5\% case over the training time. The ConRec model with the random decoder (i.e.~ConRec RD) needs more training steps but it can finally achieve the same accuracy than the initialized decoder (i.e.~ConRec).
Thus, the pretrained decoder is indeed relevant as it yields  optimal model performance with less training steps. However, given a larger number of training steps, the model can reach the same performance level -- even for low-data regimes.

This observation shows that the random decoder is not the driver for the performance gap to the SimCLR model initialization, indicating that the ConRec model learns representations in the encoder, which are better suited for segmentation tasks. ConRec's embeddings might contain more encoded information about the position of different features that are unnecessary for the contrastive task.

\section{Implication for Real World Scenario} 
\label{sec:suppl_realworld_data}

The results from the previous section indicate that pretraining the decoder has no significant effect on segmentation transfer. We test this hypothesis on a real world scenario where one model is pretrained on a reconstruction task and then finetuned on a downstream segmentation task. Therefore, we reran the official Model Genesis implementation \footnote{\scriptsize{https://github.com/MrGiovanni/ModelsGenesis}}. \citet{zhou2019models} pretrain their U-Net architecture with 3D volumes of the LUNA lung nodule dataset and then evaluate it on lung nodule segmentation. We reproduce their results and add an additional run where we initialize the decoder randomly. For the random initialization, we only finetune the model one time and for the two other variants, we finetune the models three times and report mean/std dice and IoU performance on the evaluation set.

The results in Table \ref{tab:model-genesis} show that there is no significant performance difference measurable between the fully initialized model and the model with a random decoder. This supports the findings on our benchmark datasets.

\begin{table}[h!]
\caption[Lung nodule segmentation with the Model Genesis initialization]{Results from lung nodule segmentation with the Model Genesis initialization compared to the same model with a random initialization. \emph{Full Initialization} replicates the results reported by \citet{zhou2019models} (see Fig.~7, NCS).}
\label{tab:model-genesis}
	\begin{center}
		\begin{tabular}{ ccc } 
			\toprule
			Initialization & Dice coef & IoU \\
			\midrule
			Random & 71.54 & 74.47 \\ 
			Random Decoder & 75.20 $\pm$ 1.13 & 76.94 $\pm$ 0.50 \\ 
			Full Initialization & 75.10 $\pm$ 0.48 & 76.78 $\pm$ 0.45 \\ 
			\bottomrule
		\end{tabular}
	\end{center}
\end{table}

\section{How to pretrain the encoder for downstream segmentation tasks?}
Before, we could show that there is no need to transfer learn a pretrained decoder. However, comparing the results for scenario 2 to the results for scenario 4, we find that the encoder trained with SimCLR performs significantly worse than the encoder trained with ConRec or the Reconstruction task. The SimCLR model does not even bring any advantage over a random initialization.    

\subsection{Segmentation and Classification Pretraining}

We want to assess if the limited effect of the pretrained decoder is unique to the reconstruction pretext task or if this also is an observation that occurs when transferring from one segmentation task to another segmentation task. Therefore, we present a transfer learning segmentation experiment in the following.

As the decoder does not have the primary effect on segmentation transfer learning performance, we want to investigate if it is enough to train a classification model and use that for segmentation task transfer. Obtaining classification annotation is often much easier and cheaper than obtaining fine-grained segmentation annotations. 
Therefore, the insight that pretraining with a classification model also results in good segmentation performance would save much efforts for real-world problems.

We pretrain our U-Net model on one segmentation task and then finetune the model on a different segmentation task. As a pretraining task, we use the Cityscapes dataset \citep{Cordts2016Cityscapes} with the challenge to segment all cars that appear in the image. We maximize the dice coefficient as the objective function. For pretraining, we extract patches with cars from the dataset with respective ground-truth annotations.

\begin{figure}[h!]
	\centering
	\begin{subfigure}[t]{0.15\linewidth}
		\centering
		\includegraphics[width=\linewidth]{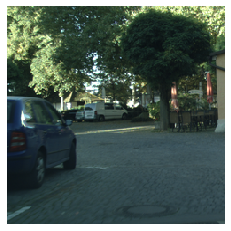}
		\includegraphics[width=\linewidth]{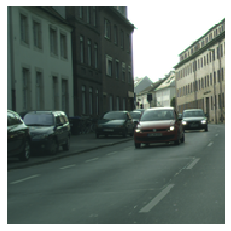}
		\subcaption{Car}
	\end{subfigure}
	\begin{subfigure}[t]{0.15\linewidth}
		\centering
		\includegraphics[width=\linewidth]{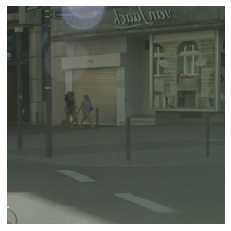}
		\includegraphics[width=\linewidth]{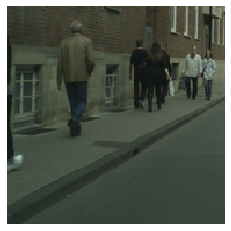}
		\subcaption{No Car}
	\end{subfigure}
	\hspace{1cm}
	\begin{subfigure}[t]{0.38\linewidth}
		\centering
		\includegraphics[width=.4\linewidth]{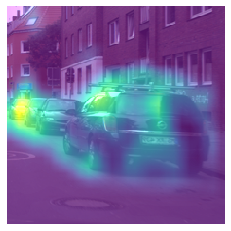}
		\includegraphics[width=.4\linewidth]{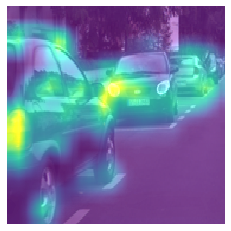}
		\includegraphics[width=.4\linewidth]{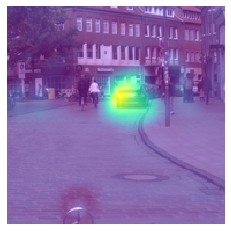}
		\includegraphics[width=.4\linewidth]{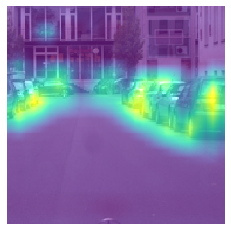}
		\subcaption{Activations of the car classification model}
		\label{fig:grad-cam-cars}
	\end{subfigure}
	\hfill
	\caption{Car classification dataset generated by cropping images from the Cityscapes dataset where either a car is present or no car is present on the image.}
	\label{fig:car-cls-dataset}
\end{figure}

To compare our segmentation pretrained model to a classification model with a comparable training task, we generate  datasets of patches showing a scene of the Cityscapes dataset and containing either a car or no car in the images. For every image of the train and validation split of the Cityscapes dataset, we decide by a coin flip if we want to generate a patch with a car or no car. Then, we take a random crop until either no pixels contain a car label (no car) or at least 20\% of the pixels have car labels (car). For model selection, we do a random 70/30 train/validation split of the patches. Samples of this dataset are provided in Fig.~\ref{fig:car-cls-dataset}. The model has to predict whether there is a car visible in the patch or not.\footnote{After pretraining the model, we used GradCam activations to validate if the model looks for cars in the image to make classification predictions -- see Fig.~\ref{fig:grad-cam-cars}.}
In both pretraining setups, we used a constant learning rate and optimized the pretraining parameters (learning rate, training epochs, weight decay) towards the best performance on the validation set.
%
%
%
%
We finetune the car segmentation model with an initialized decoder (Segmentation) and a random decoder (Segmentation-EncOnly). Furthermore, we compare these models to the model trained on the car classification task (Classification) and a random initialization (Random). We also include the results of different self-supervised approaches that have been pretrained on the whole Cityscape dataset.

\begin{table}[h!]
    \small
    \caption{Effect of the decoder when finetuning self-supervised models in comparison to finetuning a classification and a segmentation pretrained model. Displayed values indicate the dice coefficient on the evaluation set.
    }. 
    \label{tab:bus-ft}
	\centering
	\begin{tabular}{ccccc}
\toprule
Scenario & Model & pretext labels & Bus & Airplane \\
\midrule
1 (random-init)        &                  Random &                     &   75.0 $\pm$ 0.3 &         65.8 $\pm$ 0.5 \\
2  (cls-enc)        &                  SimCLR &                     &   74.5 $\pm$ 0.6 &                    68.3 $\pm$ 0.9 \\
2  (cls-enc)        &          Classification &           cars/noncars &   77.6 $\pm$ 0.9 &         68.3 $\pm$ 1.2 \\
3 (seg-enc-dec)        &          Reconstruction &                     &    76.6 $\pm$ 0.6  &                    68.7 $\pm$ 0.8 \\
3 (seg-enc-dec)        &            Segmentation &           cars/non-car &  82.3 $\pm$ 0.5  &         71.0 $\pm$ 0.7 \\
3 (seg-enc-dec)       &            ConRec &                     &   76.9 $\pm$ 1.0 &                    70.6 $\pm$ 1.7 \\
4   (seg-enc)        &  Reconstruction &                     &              76.4 $\pm$ 0.2 &                    68.2  $\pm$ 0.2 \\
4   (seg-enc)        &    Segmentation &           cars/non-car &   82.0 $\pm$ 0.6 &          70.2 $\pm$ 0.2 \\
4   (seg-enc)       &     ConRec &                     &   77.3 $\pm$ 1.1 &    70.9 $\pm$ 0.8 \\
\bottomrule
\end{tabular}
\end{table}



For evaluation, we consider two segmentation problems: bus segmentation and airplane segmentation, as introduced above. Table \ref{tab:bus-ft} shows the results after finetuning for both datasets and transfer learning scenarios. The performance gap between the segmentation-pretrained model and random initialization is larger for the bus segmentation task since the bus segmentation task is similar to the pretraining task  (i.e.~car classification). In analogy to our previous experiments, we cannot observe significant performance differences between the fully initialized model and the model with a random decoder on both datasets. We visualized the dice coefficients over the training time in Fig.~\ref{fig:dice_cityscapes_bus}. Similar to the observations for self-supervised models, the random decoder model needs more training time but then catches up to the performance of the fully initialized model. 
The classification model also yields a performance improvement compared to the random initialization but does not reach the same performance as the segmentation model. This observation indicates that although the segmentation decoder does not add much value during transfer learning, the representations in the encoder transfer better to other segmentation tasks.
By comparing the results to the self-supervised model performances on the bus segmentation dataset, we can conclude that the reconstruction based models perform similar to the car classification model but are also significantly outperformed by the model with segmentation pretraining. On the airplane dataset, the ConRec model outperforms the classification and other self-supervised models and performs on par with the segmentation model. This indicates that segmentation and classification pretraining has larger benefits for in-domain pretraining, whereas self-supervised models transfer better to different domains.


We perform an additional experiment on the Oxford IIIT Pets dataset \citep{parkhi12a} using only the dog images for pretraining. The segmentation model receives the segmentation masks of all dog images (4978 samples) and the different dog breeds (25 classes) serve as labels in the classification task. This scenario better reflects the usual multi-class classification pretraining scheme.
During finetuning, we use cat and horse images from Pascal VOC 2012 \citep{Everingham10} and compare four model initialization scenarios as shown in Table \ref{tab:pets-ablation}: 
random initialization (1); encoder pretrained with a classification (2) or segmentation (4) task; encoder-decoder pretrained with a segmentation task (3).
Similar to the airplane dataset, we use all  images from the training set where a cat/horse appears. This yields 131/68 samples for training and 119/79 samples for evaluation.

\begin{table}[h!]
\small
    \caption{Finetuning comparison between a multi-class classification and a segmentation model pretrained on dog images of Oxford Pets \citep{parkhi12a}. The values show the mean dice coefficient on the validation set of two Pascal VOC 2012 \citep{Everingham10} subsets.}
    \label{tab:pets-ablation}
	\centering
	\begin{tabular}{ccccc}
\toprule
Scenario & Model & pretext labels & VOC cat & VOC horse \\
\midrule
1 (random-init) &  Random & & 74.2 $\pm$ 1.4 & 65.0 $\pm$ 1.3 \\
2 (cls-enc)     &  Classification & dog breeds  & 76.0 $\pm$ 0.4 & 70.1 $\pm$ 0.7 \\
3 (seg-enc-dec)  &  Segmentation &  dog masks & 83.2 $\pm$ 0.2 & 79.0 $\pm$ 1.0 \\
4 (seg-enc)    &  Segmentation &  dog masks & 82.3 $\pm$ 0.2 & 77.7 $\pm$ 0.5  \\
\bottomrule
\end{tabular}
\end{table}
For both downstream segmentation tasks, a pretrained segmentation model significantly outperforms a randomly initialized model as well as a pretrained classification model.
This difference is mainly driven by the pretrained encoder -- see Scenario 2 vs.~3 and 4.
We observe a small difference between the performance of the fully initialized segmentation model and the model with a random decoder. 
Nevertheless, we observe the additional performance provided by the pretrained decoder to be small compared to the performance gain through the pretrained encoder.
These results provide additional evidence that a pretrained segmentation encoder transfers better to downstream segmentation tasks than an encoder pretrained by a classification task.

\begin{figure}[h!]
	\centering
	\includegraphics[width=\linewidth]{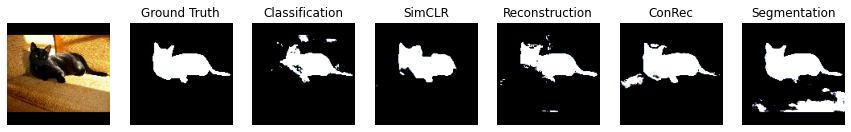}
	\includegraphics[width=\linewidth]{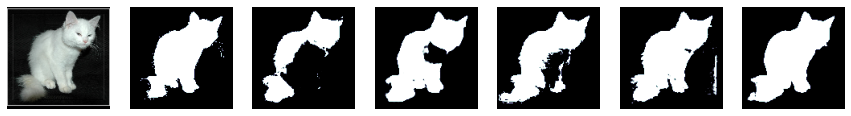}
	\includegraphics[width=\linewidth]{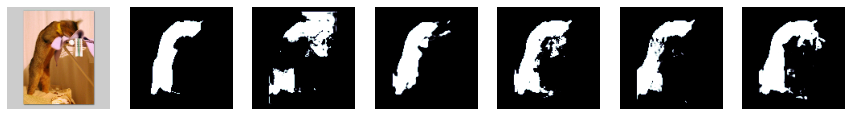}
	\caption{Segmentation predictions on the VOC cat dataset after finetuning on 131 training samples. The classification and segmentation model was pretrained on the dog images of the Oxford IIIT Pets dataset \citep{parkhi12a} whereas the self-supervised models were pretrained on the whole dataset. }
	\label{fig:qualitative-results}
\end{figure}

The performance difference between classification and segmentation pretraining can also be seen visually. Figure \ref{fig:qualitative-results} shows segmentation mask predictions after finetuning on the VOC cat dataset. 
We observe models that were pretrained on classification tasks to be less accurate on fine-grained details on the edge compared to models pretrained on segmentation/reconstruction tasks. This qualitative analysis underlines the reasoning that the classification-pretrained encoder is focussed on higher-level image characteristics.




\subsection{Encoder Output Representations}
With the above-mentioned findings, it is important to further study the representations that emerge when pretraining with a classification task, a segmentation task and self-supervised tasks. Therefore, we compare the representations of the models at the encoder output with each other. 
To compute the similarity between representations, we use \emph{Center Kernel Alignment}. As we start training neural network models from scratch and the layer activations might be of a large dimension, it is a challenging problem to compare the representations of two models.  Centered Kernel Alignment (CKA) is a method that can reliably identify correspondences between representations in networks trained from different initializations \citep{kornblith2019similarity}. CKA yields values between 0 and 1 where a larger value indicates more similarity between the representations







\begin{figure}[h!]
	\centering
	\begin{subfigure}[c]{0.4\linewidth}
		\centering
		\includegraphics[width=\textwidth]{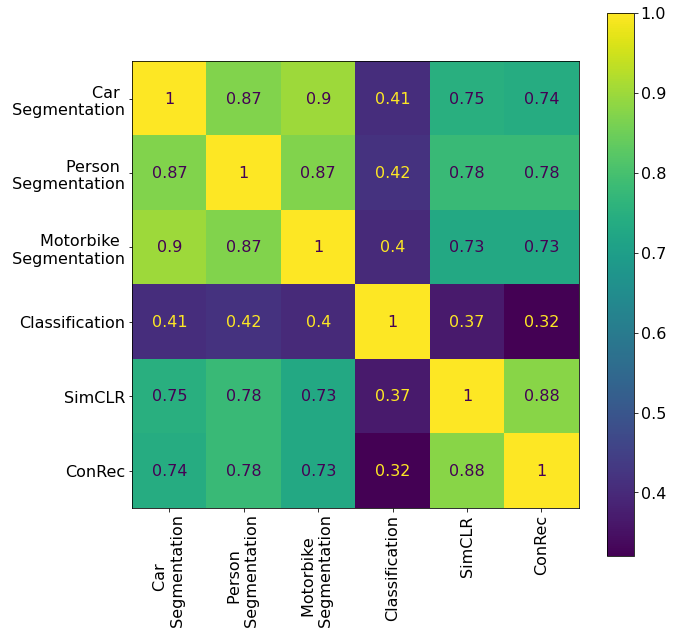}
	\end{subfigure}
	\hspace{.3cm}
	\begin{subfigure}[c]{0.22\linewidth}
		\centering
		\includegraphics[width=\linewidth]{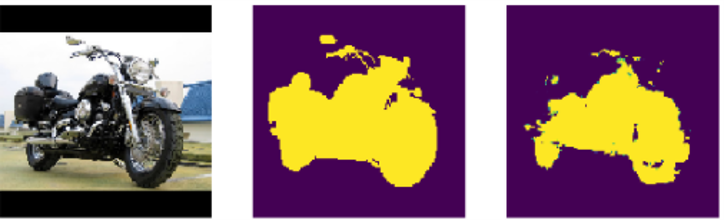}
		\includegraphics[width=\linewidth]{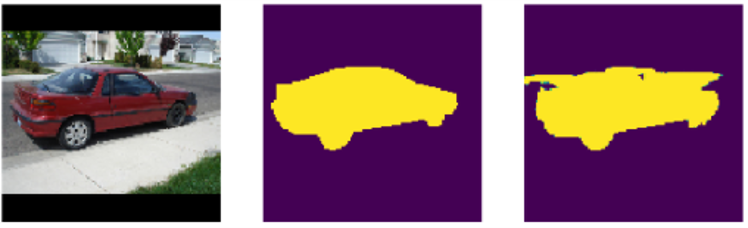}
		\includegraphics[width=\linewidth]{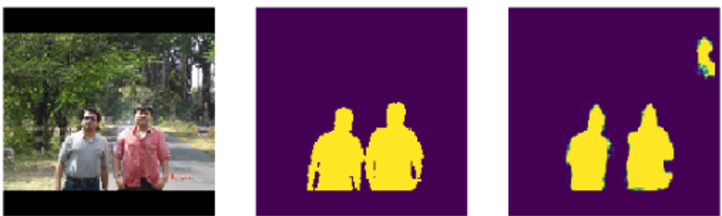}
		\subcaption{Segmentation Tasks}
		\label{fig:seg-gt-pred-examples}
	\end{subfigure}
	\hspace{.3cm}
	\begin{subfigure}[c]{0.22\linewidth}
	    \includegraphics[width=\linewidth]{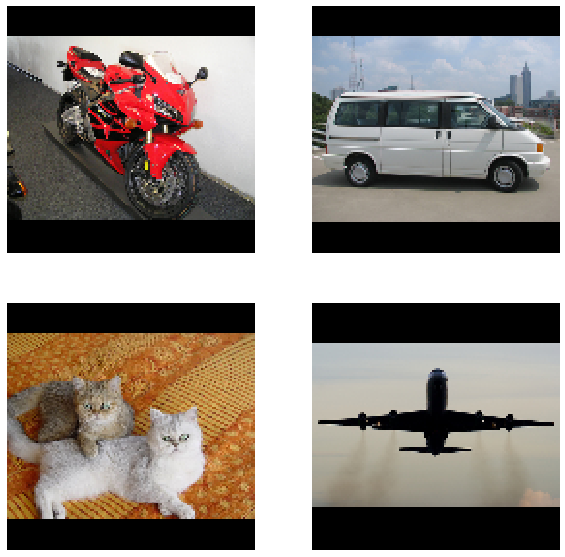}
	    \subcaption{Classification Task}
	\end{subfigure}
	\caption{Representation similarity matrix between segmentation, classification and self-supervised models on the Pascal VOC dataset. We compare the representations with Center Kernel Alignment. Larger values indicate more similarity. (a) shows examples of the three segmentation tasks with respective predictions by the learned model. (b) displays one sample for each class of the 4-class classification problem.}
	\label{fig:voc-similarity-self}
\end{figure}

We consider the Pascal VOC 2012 dataset \citep{Everingham10} as test benchmark because it contains a variety of objects and scenes. We pretrain three different segmentation models on the benchmark, each having a different segmentation target and one classification model. As segmentation targets, we choose motorbike, person, and car and only show the images where the different classes are presented to the model. We provide example images, ground truth annotations and respective model predictions
in Figure \ref{fig:seg-gt-pred-examples}.

For the classification model, we build a dataset with four different classes and the model has to distinguish between motorbikes, cars, cats, and airplanes. As this dataset contains the same images as the motorbike and car segmentation models, the models might develop similar representations during training.

After pretraining these models, we compute latent feature representations for all images from Pascal VOC, using the encoder output. We flatten the representation of size $8 \times 8 \times 512$ and then compare them pairwise between the different models with CKA. This yields a similarity matrix which is shown in Fig.~\ref{fig:voc-similarity-self}. Larger values indicate that the representations are more similar. The results show that the different segmentation models (Car Segmentation, Person Segmentation, Motorbike Segmentation) have representations that are more similar to each other compared to the representations from the classification model. 
This indicates that although the decoder may have a negligible effect for downstream segmentation tasks, the encoder learns fundamentally different representations than during a classification pretraining task. 
One simple explanation for this observation is that more spatial information is preserved at the encoder output, which is relevant for the segmentation learning task, but does not have any benefit for making classification decisions.


The self-supervised models SimCLR and ConRec have been pretrained on the training and validation split of the Pascal VOC 2012 dataset \citep{Everingham10} for 150.000 training steps. We can see that the self-supervised models all learn similar representations in comparison to the embeddings of the other models. 

We were hoping to get more insights if the improved performance of ConRec is also highlighted in the representations as building more similar representations to a downstream task. However, this analysis does not show that the self-supervised representations are closer to the representations of a downstream classification or segmentation task. The similarity to the classification and segmentation models is not significantly different between the three self-supervised models.

\section{Discussion}
We examined the benefit of using pretrained encoders and decoders for image segmentation tasks. First, we found out that there is little advantage in transferring weights from pretrained decoders: the convergence time and therefore ecological footprint gets reduced, however, the final downstream segmentation accuracy is not improved -- even in low-data regimes. Only the encoder holds valuable weights that improve the final performance. Secondly, we compared the benefit of different pretext tasks, and we propose a contrastive self-supervised approach to pretrain encoders in the absence of labels (e.g.~ConRec). We found that encoders trained on similar segmentation or reconstruction tasks perform better for downstream segmentation tasks than those trained on classification tasks. This unveils that the common practice of using ImageNet-pretrained weights for segmentation tasks is suboptimal.

We can conclude that encoders trained on different segmentation tasks are more similar to each other than to classification tasks, which we also demonstrated with a representation similarity analysis. It will be the subject of future work to validate these findings on different architectures besides U-Net. Also, our findings indicate that further experiments on large scale segmentation datasets can provide encoder weights that are more suitable for downstream segmentation tasks than the weights from the ImageNet classification task.

\bibliography{bibliography}

\begin{thebibliography}{23}
\providecommand{\natexlab}[1]{#1}
\providecommand{\url}[1]{\texttt{#1}}
\expandafter\ifx\csname urlstyle\endcsname\relax
  \providecommand{\doi}[1]{doi: #1}\else
  \providecommand{\doi}{doi: \begingroup \urlstyle{rm}\Url}\fi

\bibitem[Bachmann et~al.(2022)Bachmann, Mizrahi, Atanov, and
  Zamir]{bachmann2022multimae}
Roman Bachmann, David Mizrahi, Andrei Atanov, and Amir Zamir.
\newblock Multimae: Multi-modal multi-task masked autoencoders.
\newblock \emph{arXiv preprint arXiv:2204.01678}, 2022.

\bibitem[Bao et~al.(2021)Bao, Dong, and Wei]{bao2021beit}
Hangbo Bao, Li~Dong, and Furu Wei.
\newblock Beit: Bert pre-training of image transformers.
\newblock \emph{arXiv preprint arXiv:2106.08254}, 2021.

\bibitem[Chen et~al.(2020{\natexlab{a}})Chen, Kornblith, Norouzi, and
  Hinton]{chen2020simple}
Ting Chen, Simon Kornblith, Mohammad Norouzi, and Geoffrey~E. Hinton.
\newblock A simple framework for contrastive learning of visual
  representations.
\newblock In \emph{Proceedings of the 37th International Conference on Machine
  Learning, {ICML} 2020, 13-18 July 2020, Virtual Event}, volume 119 of
  \emph{Proceedings of Machine Learning Research}, pp.\  1597--1607. {PMLR},
  2020{\natexlab{a}}.
\newblock URL \url{http://proceedings.mlr.press/v119/chen20j.html}.

\bibitem[Chen et~al.(2020{\natexlab{b}})Chen, Kornblith, Swersky, Norouzi, and
  Hinton]{chen2020big}
Ting Chen, Simon Kornblith, Kevin Swersky, Mohammad Norouzi, and Geoffrey~E.
  Hinton.
\newblock Big self-supervised models are strong semi-supervised learners.
\newblock In Hugo Larochelle, Marc'Aurelio Ranzato, Raia Hadsell,
  Maria{-}Florina Balcan, and Hsuan{-}Tien Lin (eds.), \emph{Advances in Neural
  Information Processing Systems 33: Annual Conference on Neural Information
  Processing Systems 2020, NeurIPS 2020, December 6-12, 2020, virtual},
  2020{\natexlab{b}}.
\newblock URL
  \url{https://proceedings.neurips.cc/paper/2020/hash/fcbc95ccdd551da181207c0c1400c655-Abstract.html}.

\bibitem[Cordts et~al.(2016)Cordts, Omran, Ramos, Rehfeld, Enzweiler, Benenson,
  Franke, Roth, and Schiele]{Cordts2016Cityscapes}
Marius Cordts, Mohamed Omran, Sebastian Ramos, Timo Rehfeld, Markus Enzweiler,
  Rodrigo Benenson, Uwe Franke, Stefan Roth, and Bernt Schiele.
\newblock The cityscapes dataset for semantic urban scene understanding.
\newblock In \emph{Proc. of the IEEE Conference on Computer Vision and Pattern
  Recognition (CVPR)}, 2016.

\bibitem[Dippel et~al.(2021)Dippel, Vogler, and H{\"o}hne]{dippel2021towards}
Jonas Dippel, Steffen Vogler, and Johannes H{\"o}hne.
\newblock Towards fine-grained visual representations by combining contrastive
  learning with image reconstruction and attention-weighted pooling.
\newblock \emph{arXiv preprint arXiv:2104.04323}, 2021.

\bibitem[Everingham et~al.()Everingham, Van~Gool, Williams, Winn, and
  Zisserman]{pascal-voc-2012}
M.~Everingham, L.~Van~Gool, C.~K.~I. Williams, J.~Winn, and A.~Zisserman.
\newblock The {PASCAL} {V}isual {O}bject {C}lasses {C}hallenge 2012 {(VOC2012)}
  {R}esults.
\newblock
  http://www.pascal-network.org/challenges/VOC/voc2012/workshop/index.html.

\bibitem[Everingham et~al.(2010)Everingham, Van~Gool, Williams, Winn, and
  Zisserman]{Everingham10}
M.~Everingham, L.~Van~Gool, C.~K.~I. Williams, J.~Winn, and A.~Zisserman.
\newblock The pascal visual object classes (voc) challenge.
\newblock \emph{International Journal of Computer Vision}, 88\penalty0
  (2):\penalty0 303--338, June 2010.

\bibitem[Grill et~al.(2020)Grill, Strub, Altch{\'e}, Tallec, Richemond,
  Buchatskaya, Doersch, Avila~Pires, Guo, Gheshlaghi~Azar,
  et~al.]{grill2020bootstrap}
Jean-Bastien Grill, Florian Strub, Florent Altch{\'e}, Corentin Tallec, Pierre
  Richemond, Elena Buchatskaya, Carl Doersch, Bernardo Avila~Pires, Zhaohan
  Guo, Mohammad Gheshlaghi~Azar, et~al.
\newblock Bootstrap your own latent-a new approach to self-supervised learning.
\newblock \emph{Advances in Neural Information Processing Systems}, 33, 2020.

\bibitem[Haghighi et~al.(2020)Haghighi, Taher, Zhou, Gotway, and
  Liang]{haghighi2020learning}
Fatemeh Haghighi, Mohammad Reza~Hosseinzadeh Taher, Zongwei Zhou, Michael~B
  Gotway, and Jianming Liang.
\newblock Learning semantics-enriched representation via self-discovery,
  self-classification, and self-restoration.
\newblock In \emph{International Conference on Medical Image Computing and
  Computer-Assisted Intervention}, pp.\  137--147. Springer, 2020.

\bibitem[He et~al.(2020)He, Fan, Wu, Xie, and Girshick]{he2020momentum}
Kaiming He, Haoqi Fan, Yuxin Wu, Saining Xie, and Ross Girshick.
\newblock Momentum contrast for unsupervised visual representation learning.
\newblock In \emph{Proceedings of the IEEE/CVF Conference on Computer Vision
  and Pattern Recognition}, pp.\  9729--9738, 2020.

\bibitem[He et~al.(2021)He, Chen, Xie, Li, Doll{\'a}r, and
  Girshick]{he2021masked}
Kaiming He, Xinlei Chen, Saining Xie, Yanghao Li, Piotr Doll{\'a}r, and Ross
  Girshick.
\newblock Masked autoencoders are scalable vision learners.
\newblock \emph{arXiv preprint arXiv:2111.06377}, 2021.

\bibitem[Kornblith et~al.(2019)Kornblith, Norouzi, Lee, and
  Hinton]{kornblith2019similarity}
Simon Kornblith, Mohammad Norouzi, Honglak Lee, and Geoffrey Hinton.
\newblock Similarity of neural network representations revisited.
\newblock In \emph{International Conference on Machine Learning}, pp.\
  3519--3529. PMLR, 2019.

\bibitem[Li et~al.(2020)Li, Fan, Yuan, He, Tian, Feris, Indyk, and
  Katabi]{li2020making}
Tianhong Li, Lijie Fan, Yuan Yuan, Hao He, Yonglong Tian, Rogerio Feris, Piotr
  Indyk, and Dina Katabi.
\newblock Making contrastive learning robust to shortcuts.
\newblock \emph{arXiv preprint arXiv:2012.09962}, 2020.

\bibitem[Minaee et~al.(2021)Minaee, Boykov, Porikli, Plaza, Kehtarnavaz, and
  Terzopoulos]{minaee2021image}
Shervin Minaee, Yuri~Y Boykov, Fatih Porikli, Antonio~J Plaza, Nasser
  Kehtarnavaz, and Demetri Terzopoulos.
\newblock Image segmentation using deep learning: A survey.
\newblock \emph{IEEE Transactions on Pattern Analysis and Machine
  Intelligence}, 2021.

\bibitem[Nilsback \& Zisserman(2008)Nilsback and Zisserman]{Nilsback08}
Maria-Elena Nilsback and Andrew Zisserman.
\newblock Automated flower classification over a large number of classes.
\newblock In \emph{Indian Conference on Computer Vision, Graphics and Image
  Processing}, Dec 2008.

\bibitem[Parkhi et~al.(2012)Parkhi, Vedaldi, Zisserman, and Jawahar]{parkhi12a}
O.~M. Parkhi, A.~Vedaldi, A.~Zisserman, and C.~V. Jawahar.
\newblock Cats and dogs.
\newblock In \emph{IEEE Conference on Computer Vision and Pattern Recognition},
  2012.

\bibitem[Pathak et~al.(2016)Pathak, Krahenbuhl, Donahue, Darrell, and
  Efros]{pathak2016context}
Deepak Pathak, Philipp Krahenbuhl, Jeff Donahue, Trevor Darrell, and Alexei~A
  Efros.
\newblock Context encoders: Feature learning by inpainting.
\newblock In \emph{Proceedings of the IEEE conference on computer vision and
  pattern recognition}, pp.\  2536--2544, 2016.

\bibitem[Richemond et~al.(2020)Richemond, Grill, Altch{\'e}, Tallec, Strub,
  Brock, Smith, De, Pascanu, Piot, et~al.]{richemond2020byol}
Pierre~H Richemond, Jean-Bastien Grill, Florent Altch{\'e}, Corentin Tallec,
  Florian Strub, Andrew Brock, Samuel Smith, Soham De, Razvan Pascanu, Bilal
  Piot, et~al.
\newblock Byol works even without batch statistics.
\newblock \emph{arXiv preprint arXiv:2010.10241}, 2020.

\bibitem[Ronneberger et~al.(2015)Ronneberger, Fischer, and
  Brox]{ronneberger2015u}
Olaf Ronneberger, Philipp Fischer, and Thomas Brox.
\newblock U-net: Convolutional networks for biomedical image segmentation.
\newblock In \emph{International Conference on Medical image computing and
  computer-assisted intervention}, pp.\  234--241. Springer, 2015.

\bibitem[Zhou et~al.(2021{\natexlab{a}})Zhou, Lu, Yang, Han, and
  Yu]{zhou2021preservational}
Hong-Yu Zhou, Chixiang Lu, Sibei Yang, Xiaoguang Han, and Yizhou Yu.
\newblock Preservational learning improves self-supervised medical image models
  by reconstructing diverse contexts.
\newblock In \emph{Proceedings of the IEEE/CVF International Conference on
  Computer Vision}, pp.\  3499--3509, 2021{\natexlab{a}}.

\bibitem[Zhou et~al.(2021{\natexlab{b}})Zhou, Wei, Wang, Shen, Xie, Yuille, and
  Kong]{zhou2021ibot}
Jinghao Zhou, Chen Wei, Huiyu Wang, Wei Shen, Cihang Xie, Alan Yuille, and Tao
  Kong.
\newblock ibot: Image bert pre-training with online tokenizer.
\newblock \emph{arXiv preprint arXiv:2111.07832}, 2021{\natexlab{b}}.

\bibitem[Zhou et~al.(2019)Zhou, Sodha, Siddiquee, Feng, Tajbakhsh, Gotway, and
  Liang]{zhou2019models}
Zongwei Zhou, Vatsal Sodha, Md~Mahfuzur~Rahman Siddiquee, Ruibin Feng, Nima
  Tajbakhsh, Michael~B Gotway, and Jianming Liang.
\newblock Models genesis: Generic autodidactic models for 3d medical image
  analysis.
\newblock In \emph{International Conference on Medical Image Computing and
  Computer-Assisted Intervention}, pp.\  384--393. Springer, 2019.

\end{thebibliography}
\bibliographystyle{tmlr}

\end{document}


\title{Supplementary Material \\
\vspace{2cm}
\begin{figure}[h!]
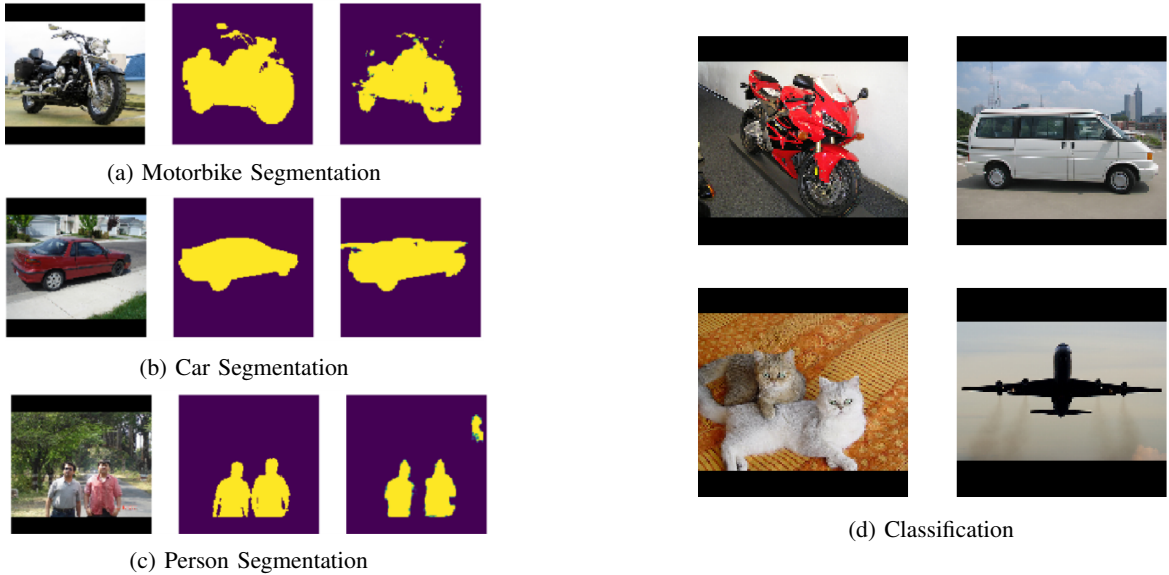

	\centering
	\begin{minipage}{.5\linewidth}
		\centering
		\begin{subfigure}[c]{0.7\linewidth}
			\centering
			\includegraphics[width=\linewidth]{figures/voc-similarity/seg/motorbike-seg.png}
			\subcaption{Motorbike Segmentation}
		\end{subfigure}
		\vspace{.1cm}
		
		\begin{subfigure}[c]{0.7\linewidth}
			\centering
			\includegraphics[width=\linewidth]{figures/voc-similarity/seg/car-seg.png}
			\subcaption{Car Segmentation}
		\end{subfigure}
		\vspace{.1cm}
		
		\begin{subfigure}[c]{0.7\linewidth}
			\centering
			\includegraphics[width=\linewidth]{figures/voc-similarity/seg/person-seg.png}
			\subcaption{Person Segmentation}
		\end{subfigure}
	\end{minipage}%
	\begin{minipage}{0.5\linewidth}
		\centering
		\begin{subfigure}[c]{0.7\linewidth}
			\centering
			\includegraphics[width=\linewidth]{figures/voc-similarity/cls-4-square.png}
			\subcaption{Classification}
		\end{subfigure}
	\end{minipage}
	\hfill
	\caption[]{Collection of Pascal VOC subset tasks used to compare initializations at the encoder output. The images on the left show example images of three different segmentation tasks with ground truth annotation and predictions by the trained model. The right side shows example images for each class concerning a 4-class classification task.}
	\label{fig:voc-models}
\end{figure}}

\maketitle
\thispagestyle{empty}